# Why Do Probabilistic Clinical Models Fail To Transport Between Sites?

*Thomas A. Lasko, Eric V. Strobl, William W. Stead*
*Vanderbilt University Medical Center*

## Abstract

The rising popularity of artificial intelligence in healthcare is highlighting the problem that a computational model achieving super-human clinical performance at its training sites may perform substantially worse at new sites. In this perspective, we present common sources for this *failure to transport*, which we divide into sources under the control of the experimenter and sources inherent to the clinical data-generating process. Of the inherent sources we look a little deeper into site-specific clinical practices that can affect the data distribution, and propose a potential solution intended to isolate the imprint of those practices on the data from the patterns of disease cause and effect that are the usual target of probabilistic clinical models.

## Main

Those of us who build prediction models from Electronic Health Record (EHR) data commonly find that a model that works well at its original site doesn't work nearly as well at some other site[1-4]. (By *site* we mean a location in time and space, so the failure could be at the same institution, but a later time). When the new site's data is included in training and test sets, performance improves to match that of the original[3], but then performance at a third site can be back to nearly random. While this *failure to transport* is quite frustrating, we argue that we should expect this sort of behavior from all probabilistic clinical models, whether they are supervised predictive models or unsupervised discovery models, even if we are meticulous in their training.

Experimental errors, improper analysis, and failure to document experimental details can lead to a *failure to replicate*, in which an intended identical experiment performed at a new site produces conflicting results[5-8]. These issues can be difficult to avoid[9-11], and the problem is exacerbated by a weak culture of replication[12]. But while failure to transport could be considered an instance of failure to replicate[13-16], we see it as a sufficiently distinct phenomenon to warrant specific attention.

Failure to transport was recognized as a problem by the earliest medical AI pioneers. In 1961, Homer Warner implemented the first probabilistic model of symptoms and disease[17] using a Naïve Bayes method[18] with local conditional probabilities, which then failed on external data[19]. A follow-up by Bruce and Yarnall[19] using data from three sites noted similar failure to transport between sites, which they attributed to differences in conditional probabilities. A decade later,

Alvan Feinstein argued that the very idea of probabilistic diagnosis was fatally flawed, on the grounds that the observational accuracy, the prevalence, and even the definitions of collected clinical observations varied across sites[20]. A decade after that, in the inaugural issue of *Medical Decision Making*, Tim de Dombal doubted ever being able to design a probabilistic diagnosis engine with data from one site that worked at others, because large-scale surveys demonstrated that disease prevalence and presentation vary dramatically across locations[21].

All probabilistic transport failure can be attributed to differences in the multivariate distribution of the training dataset vs. the application dataset, because the dataset is the sole means by which site-specific phenomena communicate with the model. (We use *application dataset* to refer to the data with which the model will be used in practice, as opposed to the *training* and *test sets* that are used during development.) The difference could include slight mismatches in the univariate distribution of a variable, or more radical differences in the dependencies between many variables.

In this perspective, we consider various sources of these differences, dividing them into *Experimental Sources* that can be minimized by the experimental configuration, and *Inherent Sources* that, because they are internal to the way the clinical data are generated, are not so easy to avoid (Table 1). (We use the term *unstable* for distributions that vary between sites due to either experimental or inherent sources[22–24].) Of the inherent sources, we will discuss in some depth the problem of site-specific clinical workflow processes that are difficult to account for.

## Experimental Sources

Some distributional differences between training and application datasets can be minimized by sufficient attention to the training pipeline. We call the sources of these differences *experimental* in the broad sense that training any model is a computational experiment, regardless of whether the model is supervised or unsupervised, whether it uses observational or interventional data, or whether it is addressing a hypothesis-driven or discovery-based question. Experimental sources of instability are quite common, and they can be difficult to recognize.

### Model Overfitting

An overfit model performs well on training data but poorly on test data, even though the two datasets are drawn from the *same* underlying distribution[25]. Despite the fact that there should be no distributional difference between the two datasets, an overfit model has come to rely in part on patterns that are present *only by chance* in the training set, and which are necessarily different in application data. Getting a model to perform well while avoiding overfitting is the central task of machine learning[25], and we continue to learn surprising things about it[26–29]. A noticeable subset of failure-to-transport results, including those getting recent attention[13], is really just unrecognized overfitting[30].

Table 1: Potential Sources of Model Transport Failure

| **Experimental Sources** |
| --- |
|     Model Overfitting |
|     Information Leaks During Training |
|     Different Variable Definitions in Application Data |
|     Application to the Wrong Question |
| **Inherent Sources** |
|     Application Data with Different Causal Prevalence |
|     Presence of Site-Specific Processes |

## Information Leaks During Training

The unintended presence of information in a training dataset that would not be present at application time is an *information leak*[31,32], a machine-learning equivalent to inadvertent unblinding in a clinical trial[33]. For example, mostly-positive and mostly-negative cases may be collected from separate sources, then Patient IDs assigned sequentially by source. Given this data for training, a model can easily learn that earlier Patient IDs are more likely to be positive[31].

A common leak is the use of positive instances constructed from, say, the time of hospital admission to the predicted event of interest (perhaps the onset of sepsis), and negative instances constructed from the full length of the admission (because there was no sepsis). In this case, variables such as time since admission or the presence of typically near-discharge events (such as weaning from a ventilator) leak information about the probability of a positive label.

Any information generated after the intended moment of application can leak information about the label. Often, this information leaks from times after the label was known, but more subtle leaks can occur from earlier times[32]. For example, information about treating the condition (perhaps starting antibiotics for sepsis) can make its way into training instances, leaking information about the condition's presence. But other indicators such as fever or tachycardia can originate from the period before sepsis is formally labeled, but clinically obvious, and a model predicting sepsis at that point would be less clinically useful[32].

In addition to information leaking from the future, information can also leak between the training, test, and validation sets[31]. These leaks can be easy to miss, as in the well-known CheXNet paper[34], where researchers originally split the dataset by image, rather than by patient, leaking patient-level information between training and test sets[35]. (Although even after the leak was corrected, this impressive model still failed to transport to other sites[36–38].)

## Different Variable Definitions in Application Data

If an application dataset defines a prediction target differently from the training data, model performance will suffer. This may seem obvious, but its role in failure to transport is easily missed, as happened[39] when the performance of EHR vendor Epic's internal sepsis prediction model dropped at least in part due to a more careful definition of the sepsis label in the application dataset[40].

Similarly, if semantically equivalent data variables are encoded with different identifiers in a new dataset, then the performance of a model that relies on those identifiers will suffer[41,42]. This may also seem obvious, but it can be a subtle problem, because, for example, different institutions that both use LOINC codes[43,44] to identify laboratory test results may actually use *different* LOINC codes for clinically equivalent tests[45]. And, of course, test results can be reported in different units (and possibly mislabeled) even within the same dataset[46].

A definitional mismatch can cause the model to see what appear to be distribution differences in the affected variables, when in reality the model is seeing different variables with the same name.

*Application to the Wrong Question*

A model trained to answer *question A* is not likely to be as accurate if we try to use it to answer *question B*. This is yet another obvious statement, but again the problem can be subtle and easy to misdiagnose, especially if the questions are related.

This has happened multiple times with a 1997 model that predicts the mortality of pneumonia in hospitalized patients[47]. Researchers trained the model to answer question A, for which they had data: *How likely is this patient to survive when given usual inpatient care?* They then applied it to answering question B, which is what they really wanted to know: *How likely is this patient to survive if sent home **without** inpatient care?* The original researchers explicitly stated that they were assuming that the answer to the two questions would be similar in patients with low probability of inpatient mortality.

Unfortunately, subsequent interpretations of the work appear to have missed the explicit assumption and misdiagnosed the source of resulting problems. A 2015 evaluation[48] of an interpretable learning method identified cases where the assumption was violated, such as for patients with asthma. Pneumonia patients with asthma were more likely than other pneumonia patients to survive with usual inpatient care because they were more likely to be immediately admitted to the ICU. But they are of course much *less* likely to survive if sent home *without* inpatient care. The 2015 work shows the clear value of interpretable models, because it highlighted this problem. But rather than pointing out serious assumption violations and the application to the wrong question, the researchers' next step was to patch the model by manually changing the weights for the *asthma* term (and any others that seemed counterintuitive). Later authors, citing only the 2015 work, claimed that the error was an unintended consequence of machine learning because the model had learned subtle patterns and was missing vital context[49]. Others recognized that data distribution appeared distorted[50,51] (which it was, relative to question B), but that wasn't the core problem. The core problem was that the model was answering a different question than its application users were asking[52]. Alternatively, we could say that the users were asking a causal, counterfactual question that the (non-causal) predictive model was not designed to answer: "What would happen if we intervened by sending the patient home instead of the usual practice of admitting to the hospital?"

## Inherent Sources

Application distributions can differ from training distributions because some real phenomenon of interest affects them differently (as opposed to overfitting, in which the difference is only due to random variation). Using a model under these conditions is an Out-of-Distribution (OOD) application[53–56]. Optimizing for OOD performance is an interesting and growing research direction[56–66].

The need to consider OOD performance in clinical prediction models arises from the fact that, as the AI pioneers observed, clinical data distributions actually do change between sites for reasons inherent to the data-generating mechanism, a phenomenon now known as *dataset shift*[54] or *distribution shift*[57]. As models have grown in their power to use complex distributional patterns, so has their need to adapt to this shift.

## Application Data with Different Causal Prevalence

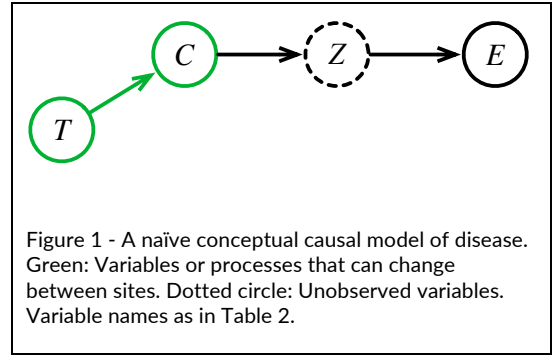

Figure 1 - A naïve conceptual causal model of disease. Green: Variables or processes that can change between sites. Dotted circle: Unobserved variables. Variable names as in Table 2.

The most obvious distributional difference between sites is the underlying disease prevalence, which can vary drastically. If the prevalence of the predicted outcome differs between the training data and the application data, this can affect the performance of a model, especially its calibration[67,68]. Fortunately, differences in prevalence are easy to accommodate, at least in simple linear models[67,69,70].

However, while we usually speak conceptually about prevalence of *disease*, what actually varies is the prevalence of its *causes*. Consider a naïve probabilistic causal model of disease (Figure 1), where $Z$ is the core conceptual problem, with its causes $C$, and effects $E$. When the prevalence of causes varies with the site $T$, that will affect the downstream prevalence of the disease and its effects.

The path $C \to Z \to E$ is an abstraction representing a large network of causes and effects; we choose a node $Z$ as our condition of interest, which defines upstream nodes as causes, and downstream nodes as effects.

For example, the core conceptual problem of Acute Myocardial Infarction (AMI, or heart attack) is the sudden death of heart muscle. There are many causes of AMI, of which the most immediate is an abrupt reduction in blood flow to the heart muscle. Causes further upstream in the network may include arteries narrowed by plaque, a sudden rupture of that plaque, a blood clot abruptly blocking an artery, or vasospasm stimulated by mediators released from platelets.

So far, none of the causes should inherently vary between sites. But even further upstream are causes related to genetics, diet, exercise, and environmental or medication exposures, all of which do vary with geographic location, local patient populations, local practice patterns, or time. That variation will eventually affect the downstream prevalence of AMI in the population, as well as that of any downstream effects, such as chest pain, shortness of breath, release into the bloodstream of intracellular cardiac enzymes, unconsciousness, or death.

Sometimes, a condition of interest is really the union of a set of subtypes, within which some are easier to predict than others[38,71,72]. In this case, a change in the distribution of subtypes will affect the performance of the model. Performance can even improve on an external dataset if it contains a larger proportion of an easy subtype. This fact can be used nefariously to construct test sets that dramatically increase the apparent performance of a model by including a high prevalence of an easily predicted subtype (such as an easy negative subtype).

## Presence of Site-Specific Processes

With some thought, we can identify additional inherent sources of instability that are not represented in Figure 1. To understand these, note that in general, the variables in the $C \to Z \to E$ *cause-effect network* are not directly recorded in clinical practice[73,74], largely because (as with heart muscle death) they are not directly observable under typical clinical conditions. Instead, the related *observational* variables $O$ (such as laboratory test results, clinical images, medication records, billing codes, and narrative clinical text) are observed and recorded using site-specific processes (Figure 2). While these are intended to reliably reflect the latent variables in the cause-effect network, in practice they cover only a subset of those variables, and the observations depend on case mix, practice patterns, specific instruments, reagents, and personnel, as well as financial incentives[71,75,76]. We might expect this dependence to be standardized across sites, but it turns out to be an important source of distribution shift, even for observations as straightforward clinical laboratory tests, which are affected by spectrum bias[72], test ordering patterns[77], variation in measurement[78], and variation in reference values[79].

Table 2: Variable Names

| |
|---|
| $T$: Data Collection Site (in time and space) |
| $C$: (Latent) Disease causes or risk factors. |
| $Z$: (Latent) Core disease characteristics. |
| $E$: (Latent) Downstream effects. |
| $Y$: Binary label or continuous target |
| $O$: Observed variables |
| $S$: Selection variables |

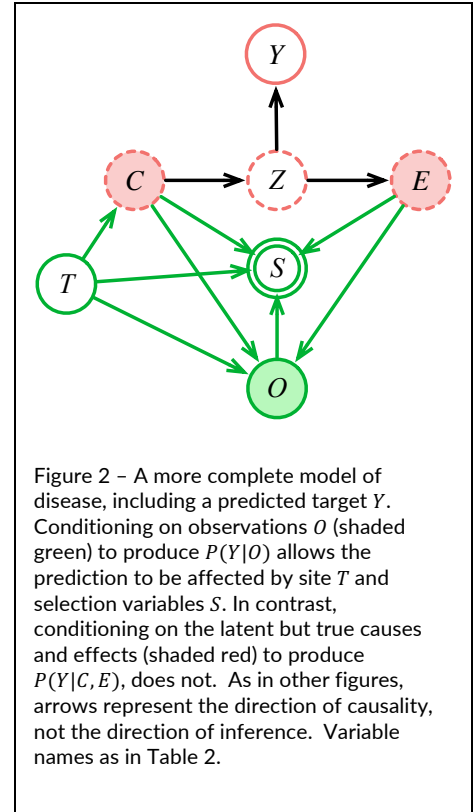

Figure 2 – A more complete model of disease, including a predicted target $Y$. Conditioning on observations $O$ (shaded green) to produce $P(Y|O)$ allows the prediction to be affected by site $T$ and selection variables $S$. In contrast, conditioning on the latent but true causes and effects (shaded red) to produce $P(Y|C,E)$, does not. As in other figures, arrows represent the direction of causality, not the direction of inference. Variable names as in Table 2.

Instability arises not only from *how* a variable is observed, but also *which* variables and *when*[66,77]. The *which* and *when* decisions are generally made by expert recognition of a developing clinical picture, and can depend on any other variable or process at the time of the decision. These include pathophysiologic phenomena, but also practical phenomena such as the patient being discovered unresponsive in public, brought to the hospital by ambulance, being assigned to a particular clinical team, or other factors that may or may not be recorded in the EHR. They may also include the upstream decision by the patient to present for care.

Decisions to observe are represented in our model by a set of selection variables $S$ (Figure 2)[80]. The variable $S_o \in S$ represents the decision of *when* to record (record when $S_o = 1$, not when $S_o = 0$), and each $S_{O_i} \subseteq S$ is a set of variables that collectively represent the decision of *which* $O_i$ to record (record $O_i$ when all elements of $S_{O_i} = 1$). Any decision in $S$ may depend on $T$, $C$, $E$, or $O$. The *when* and *which* decisions together produce what has been described as "selection bias on selection bias"[80]. (For simplicity, we omit a discrete time index $t$ in the graph and the notation. But the notation can be extended for all variables so that, for example, $S_o$ becomes $S_o[t]$, meaning $S_o$ at the time $t$.)

In clinical practice, many measurements are prompted by the suspicion that a relevant observational variable $O_i$ may be outside of its healthy range, which means that the distribution $P(O_i|S_{O_i} = 1)$ is different from $P(O_i|S_{O_i} = 0)$, also known as *informative missingness*[81] or *informative presence*[82]. The decision to observe is subject to disagreement about what is relevant (even between clinicians of the same specialty trying to answer the same clinical question about the same patient[83–86]), and can become subject to feedback loops if the decisions themselves are used as predictors in a model[81,87]. Given this variation in *how, when,* and *which* observations are made, it is difficult to see how *any* of the observational variables could be stable across sites.

Typically, we want to estimate $P(Y|O)$. But conditioning on observations $O$ does not block the influence of the site $T$, which is the root cause of all instability (Figure 2), and a sufficiently powerful model can estimate that influence, given enough data. This can happen even if the observations in $O$ are only the pixels of a radiographic image[88]. A powerful model can learn what it means clinically that a chest X-ray was done on a particular machine, that a laboratory test was done on a weekend, or that it was ordered by a given physician. In a multi-site dataset, the identity of the originating site can easily be inferred from site-specific dependencies and exploited for prediction[37].

Perhaps because the model is exploiting unanticipated dependencies in the data that don't match causal pathophysiologic pathways, these dependencies have been called *shortcuts*[89], and considered cheating. But blaming the model isn't a productive research direction. The algorithm doesn't know what we intend for it to learn. It can't tell the difference between pathophysiologic and process-related patterns. All it knows is that the task is to take $O \in R^n$ and predict $Y \in R$ or $Y \in \{0,1\}$, and it uses all available patterns to do that.

Moreover, human experts also exploit process-related information. A radiologist doesn't just look at the film to identify signs of disease, she also wants to know why the film was ordered, who ordered it, and what was going on with the patient at the time. A pathologist doesn't only look at the slide under the microscope, he also wants to know the location on the body where the biopsy was taken, and what was the clinical scenario that led to ordering it. If a computational model can infer such things from the input data, there is no reason why it shouldn't use them to improve its performance.

## Potential Solutions

Experimental sources of instability have known solutions that can take effort to implement, but these are at least under the control of the experimenter[14,16,90]. Minimizing the effects of inherent sources is a harder problem, because they are actually part of the data-generating mechanism[54,91,89,92,93]. We are unaware of any experiments to quantify the relative prevalence and magnitude of experimental vs. inherent sources of instability, although we argue above that inherent sources are likely to drive many transport failures. If this turns out to be true, how could we minimize their impact?

An obvious solution is to somehow identify and exclude the inherent unstable patterns from the model. There are powerful and sophisticated methods to do this, whether the unstable information resides in nodes, edges, or a combination[60,51,56,65,66,94–96]. A related approach is to use data from multiple sites during training, under leave-one-site-out cross validation, so the model identifies only stable features to begin with[2]. These all can be effective methods in specific circumstances. However, we might expect that removing *any* non-redundant information will decrease a model's performance, inevitably trading performance for stability[56]. If unstable patterns are as ubiquitous as expected in clinical data, removing them could have a disastrous effect.

Is it possible to keep the information contained in unstable patterns, but minimize their impact on transportability? One potential direction is to notice that the task of training a stable model reduces to blocking the influence of the site $T$ on the estimate of disease variables $Z$ and $Y$. There are a host of ways of doing this, including carefully collecting prospectively randomized data, or specifying an appropriate set of conditioning variables using prior knowledge to block the effects in a particular dataset[66]. There are also measurement error models that can correct for measurement differences between sites[97,98]. These approaches can be effective, but properly specifying the appropriate conditioning set or measurement error distribution remains challenging in this setting. Instead, we would like an approach that could be more easily applied to untamed EHR data.

In general, we can block the influence of the site $T$ on the estimate of disease variables $Z$ and $Y$ by conditioning on the latent cause and effect variables $C$ and $E$, because $Z, Y \perp\!\!\!\perp T \mid C, E$ (Figure 2). However, conditioning on latent variables is difficult; to address this difficulty, we consider separating our model into a *Process Model* that estimates the latent variables, and a *Disease Model* that uses them for prediction.

We expect that both models would need to be learned from data. The Process Model must infer the site-specific relationships $P(C, E|O, T, S)$, where $S$ is given implicitly by the missing elements of $O$. It represents the imprint on the data of local implementations of care processes, and must be trained for each site. The Disease Model would infer the stable relationships $P(Z|C, E)$, and $P(Y|Z)$ if a specific label or target is required (Figure 3). It represents how disease behaves; it generalizes across all sites, and could even be trained using estimates $\{C, Z, E, Y\}$ pooled from multiple sites.

To achieve stability under this arrangement, the Process Model must infer sufficiently accurate point estimates of the latent variables, or else information about $T$ can leak through the conditioning. Given enough information in $O$, such as by a sufficiently large number of variables in $O$, acceptable accuracy should be achievable[99,100]. We know of no implementations of this strategy, but some promising initial steps have been made.

First, de Fauw and colleagues[101] found that a model of retinal disease from 3-dimensional optical coherence tomography images ported poorly between different types of scanners, with an error rate of 46% on external data for predicting specialty referral (vs. 5.5% on the internal test set). To address the instability, they created a Process Model that produced a tissue segmentation $P(C, E|O, T)$, from the raw image pixels and a Disease Model $P(Z, Y|C|E)$ that recognized retinal pathology from the segmentation and made a referral

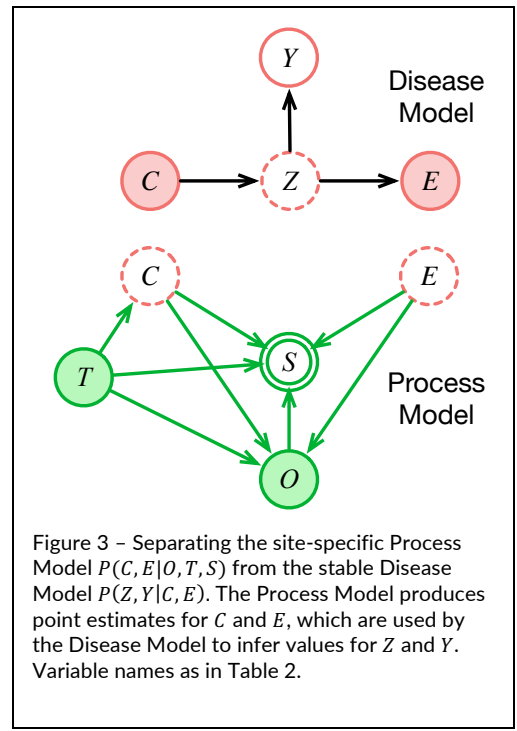

Figure 3 – Separating the site-specific Process Model $P(C, E|O, T, S)$ from the stable Disease Model $P(Z, Y|C, E)$. The Process Model produces point estimates for $C$ and $E$, which are used by the Disease Model to infer values for $Z$ and $Y$. Variable names as in Table 2.

recommendation. Retraining only the Process Model on the new scanner's data dramatically reduced the external error rate to 3%. This worked because the investigators knew what the latent variables were (the physical structure of tissue layers), and were able to label them for training the Process Model. But the results demonstrate the promise of concentrating the instability into the Process Model.

Next, Lasko and Mesa[102] used probabilistic independence to infer unsupervised data signatures that represent 2000 latent variables in the $C \rightarrow Z \rightarrow E$ network from a large EHR dataset. Then they separately trained a supervised Disease Model using the estimated values of those variables to predict liver transplant 10 years in the future. The top predictors in the Disease Model corresponded *in correct rank order* to the leading causes of hepatocellular carcinoma, suggesting that the method had accurately estimated causal latent variables. Strobl and Lasko[103] later proved theoretically that the latent variables learned by this approach do, in fact, correspond to root causes of disease (meaning the furthest upstream variables in C or Z with unique causal effects on $Y$), and can identify those causes specific to each patient case. These results demonstrate the promise of estimating and then conditioning on latent disease variables, even when they are not known in advance. The approach has since been extended to accommodate heteroskedasticity[104] and latent confounders[105].

But is it actually worth the effort to separate one model into two? If we must retrain a Process Model at each site anyway, why not just retrain the whole thing? For example, why not start with a dataset from just a few sites, using a site identifier as an input variable? Additional sites could be added to the model by simply continuing the training using the new site's data, with its new site identifier. The question recalls de Dombal's note that a model using global probabilities performed much better at each data-contributing site than cross-site probabilities did, though not quite as well as same-site probabilities[21]. A lighter variation on this could be to develop a single foundational model, which each site would fine-tune or update separately with local data[3,69,106–108]. These approaches can be quite practical and effective.

The largest benefit we see to our proposal of isolating the two types of patterns is that while the updated or cumulative model solutions may address the transportability problem, they miss out on what could be substantial advances that exploit the natural division between process relationships and disease relationships.

First, we see the Disease Model as representing what we actually want to know about health and disease. It is described using the same relationships that guide clinical thinking, that clinicians learn in the classroom phase of medical school. It is stable, transferrable knowledge that can be directly shared between institutions, forming the core of a learning healthcare system. It is the representation that abstracts away all site-specific information.

And second, we see the Process Model as representing the site-specific processes that clinicians learn in their clerkships and residency, by which patients are diagnosed, monitored, and treated. It gives a window onto what we want to know about healthcare delivery: What are our care processes? How do they differ between institutions? How do they evolve over time? What are their inefficiencies and conflicts? Those questions are probably best answered by direct observation, but the processes involved do leave an imprint on the data record, which when isolated could provide clues or signals about the processes. This idea recalls Hripcsak and Albers[76] from a decade ago, who insightfully described the EHR as an artifact of the recording process, rather than a direct record of disease. They called for studying the EHR as a phenomenon of its own, "deconvolving" the actual patient state from what is recorded. Their call maps directly onto learning and analyzing Process Models.

## Summary

It should not surprise us when high-performance probabilistic clinical models fail to transport to new sites. The failure is directly caused by differences in the multivariate distribution between the training data and the application data. Some of these differences are due to controllable experimental factors, such as overfitting, information leaks, data definitions, and misuse, but others are inherent in site-specific data-generating mechanisms, and much more difficult to avoid. How best to minimize the inherent factors is an open question, as is the relative impact of experimental vs inherent sources in transport failure. However, we argue that simply removing unstable patterns and variables from clinical datasets is unlikely to succeed, because current medical practice renders unstable nearly all recorded observations and the dependencies that relate them to disease attributes of interest. Instead, we envision a path to embracing the instability and exploiting all information in a record to maximize performance, by training first a Process Model that uses the unstable site-specific information to infer latent cause and effect variables, and then a Disease Model that uses those latent variables to make stable inferences about a patient's clinical state.  While there are only early indicators of potential merit in the approach, we believe that the perspective presents a promising research direction for disentangling disease from clinical processes.

## Acknowledgements

This work was funded in part by grants from the National Library of Medicine (LM013807), the National Cancer Institute (CA253923), and the National Institute of Arthritis and Musculoskeletal and Skin Diseases (AR076516).

## Competing Interests

The authors declare that there are no competing interests.

## Author Contribution

TAL conceived the original direction, with substantial scientific refinement in discussion with EVS and WWS. TAL wrote the initial manuscript draft, with critical revision by EVS and WWS. All authors accept responsibility for the final article.

## Data Availability

No data were used or generated by this work.